\documentclass{article}

\usepackage[final]{neurips_2025}

\usepackage[utf8]{inputenc} 
\usepackage[T1]{fontenc}    
\usepackage{hyperref}       
\usepackage{url}            
\usepackage{booktabs}       
\usepackage{amsfonts}       
\usepackage{nicefrac}       
\usepackage{microtype}      
\usepackage{xcolor}         

\usepackage{graphicx}
\usepackage{amsmath}

\title{CGS-GAN: 3D Consistent Gaussian Splatting GANs for High Resolution Human Head Synthesis}

\author{%
  Florian Barthel \\
  Fraunhofer HHI \\
  HU Berlin \\
  \And
  Wieland\\
  \textbf{Morgenstern} \\
  Fraunhofer HHI \\
  \And
  Paul Hinzer \\
  Fraunhofer HHI \\
  \And
  Anna Hilsmann \\
  Fraunhofer HHI \\
  \And
  Peter Eisert \\
  Fraunhofer HHI \\
  HU Berlin\\
}

\begin{document}


\maketitle
\begin{figure}[h!]
    \centering
    \includegraphics[width=1\linewidth]{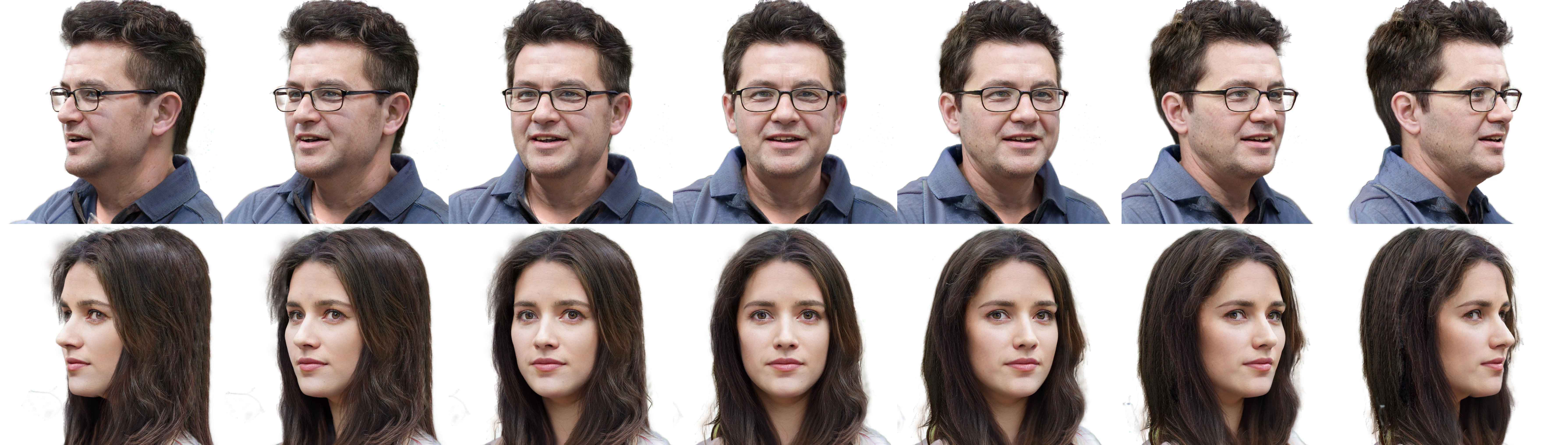}
    \caption{Example renderings at $1024^2$ resolution produced by our 3D \textit{Consistent Gaussian splatting GAN} (CGS-GAN). Unlike prior methods that recompute the scene for each individual view to ensure high quality, our method is capable of preserving quality while synthesizing a fully 3D consistent scene that can be ported into explicit 3D settings like game engines or VR environments.}
    \label{fig:teaser}
\end{figure}

\begin{abstract}
Recently, 3D GANs based on 3D Gaussian splatting have been proposed for high quality synthesis of human heads. However, existing methods stabilize training and enhance rendering quality from steep viewpoints by conditioning the random latent vector on the current camera position. This compromises 3D consistency, as we observe significant identity changes when re-synthesizing the 3D head with each camera shift. Conversely, fixing the camera to a single viewpoint yields high-quality renderings for that perspective but results in poor performance for novel views. Removing view-conditioning typically destabilizes GAN training, often causing the training to collapse. In response to these challenges, we introduce CGS-GAN, a novel 3D Gaussian Splatting GAN framework that enables stable training and high-quality 3D-consistent synthesis of human heads without relying on view-conditioning. To ensure training stability, we introduce a multi-view regularization technique that enhances generator convergence with minimal computational overhead. Additionally, we adapt the conditional loss used in existing 3D Gaussian splatting GANs and propose a generator architecture designed to not only stabilize training but also facilitate efficient rendering and straightforward scaling, enabling output resolutions up to $2048^2$. To evaluate the capabilities of CGS-GAN, we curate a new dataset derived from FFHQ. This dataset enables very high resolutions, focuses on larger portions of the human head, reduces view-dependent artifacts for improved 3D consistency, and excludes images where subjects are obscured by hands or other objects. As a result, our approach achieves very high rendering quality, supported by competitive FID scores, while ensuring consistent 3D scene generation.
Check our our project page here: \url{https://fraunhoferhhi.github.io/cgs-gan/}
\end{abstract}

\section{Introduction}
\label{sec:intro}

Synthesizing high-quality 3D human heads remains a major challenge in computer vision with wide applicability in film and gaming industries. However, this often requires expensive 3D capturing studios or high manual labor. For this reason, there is a high demand for automating and simplifying the synthesis process while maintaining high rendering quality. 3D GANs trained on 2D images with known camera poses have emerged as a powerful framework, enabling both generation and manipulation of facial geometry and appearance. Earlier approaches such as 2.5D GANs encode camera viewpoints directly into the generator features, allowing for multi-view synthesis but often lacking consistency across viewpoints.\cite{Shoshan_2021_ICCV,styleflow,Brehm2022-dx,nguyen2019hologan}. 3D-aware GANs based on Neural Radiance Fields \cite{barron2021mipnerf,mildenhall2020nerf} improve view disentanglement by rendering from an intermediate 3D volume, but suffer from slow inference and limited resolution scalability \cite{xue2022giraffe,Chan2021,chanmonteiro2020pi-GAN,an2023panohead,NEURIPS2020_e92e1b47}. More recently, 3D Gaussian Splatting (3DGS) \cite{kerbl3Dgaussians} has been integrated into generative models, providing explicit scene representations with real-time differentiable rendering capabilities. Methods like GSGAN \cite{hyun2024gsgan} and GGHead \cite{kirschstein2024gghead} demonstrate that high-quality 3D Gaussian head synthesis is feasible within GAN frameworks. However, they rely on view-conditioning, where camera parameters are fused with latent codes to guide synthesis. While this aids training convergence, it can lead to view-dependent variations in geometry and identity.

To address this, we introduce CGS-GAN, a novel 3DGS-based generative framework that achieves stable training and 3D-consistent synthesis of human heads without the need for view-conditioning. Unlike existing methods 
, CGS-GAN achieves very high rendering quality from any camera pose without re-synthesizing the 3D scene for each view. This is made possible through a lightweight multi-view regularization strategy that stabilizes the training and encourages geometric consistency with minimal computational overhead. To support high-resolution synthesis and scalability, we propose an efficient generator architecture that minimizes GPU memory usage while enabling output resolutions up to $2048^2$. Additionally, we apply random background augmentation to mitigate hole artifacts. To underline the capabilities of our new GAN framework, we curate a high-quality dataset derived from FFHQ, including entire heads by recropping from 4K sources, excluding occluded faces, and reduce view-specific biases. Our key contributions are:

\begin{itemize}
\setlength{\itemindent}{-1em}
\item We propose a novel 3DGS-based GAN framework that synthesizes consistent 3D human heads.
\item We introduce a multi-view regularization that stabilizes training and promotes 3D consistency.
\item We propose a memory-efficient generator design supporting scalable, high-resolution synthesis.
\item We apply a random background augmentation method that reduces common artifacts.
\item We curate a high-quality FFHQ-based dataset for high-fidelity, 3D consistent head modeling.
\end{itemize}
\section{Related Work}
\label{sec:related_work}

Generative adversarial networks (GANs) were first introduced by Ian Goodfellow in 2014 \cite{goodfellow2014generative}. Ever since, GANs have received high attention for their real-time photorealistic synthesis capabilities. In its core, GANs are constructed as a non-cooperative game between a generator that synthesizes images from random latent vectors and a discriminator that differentiates between fake images and real images from a training dataset.
The training pipeline is differentiable so that the generator can be optimized to produce outputs that the discriminator confuses with real images.
As both networks improve during training, the generator converges towards the training data distribution. As synthesizing from random latent vectors creates only limited control, conditional GANs (cGANs) \cite{condGAN} have been introduced as an extension to GANs, allowing annotations to be incorporated into the GAN training framework. 

\subsection{3D GANs}
While early 3D GANs were based on conditional 2.5D GANs \cite{Shoshan_2021_ICCV,styleflow,Brehm2022-dx,nguyen2019hologan} or generating voxel grids \cite{shape17,NIPS2016_44f683a8}, there have been drastic improvements since the introduction to \textit{Neural Radiance Field (NeRF)} \cite{barron2021mipnerf,mildenhall2020nerf}. NeRFs provide a differentiable renderer that allows for high quality synthesis, by querying a small \textit{Multi-layer Perceptron} MLP to decode color and opacity during ray tracing. By designing a generator that directly produces the weights of a NeRF (MLP), PI-GAN \cite{chanmonteiro2020pi-GAN} and GRAF \cite{NEURIPS2020_e92e1b47} were the first works to include a NeRF renderer in a 3D GAN framework. Although the rendering quality was promising, these approaches were slow to render and limited in output resolution. Addressing this, EG3D \cite{Chan2021} proposes an efficient 3D GAN architecture that introduces a lightweight tri-plane representation, which allows 2D features to describe a 3D volume that can be rendered using a very small NeRF MLP. This architecture considerably sped up the training and inference. Nevertheless, EG3D still relied on 2D super-resolution to render in high resolutions after the 3D rendering, which harms the 3D consistency. To achieve better 3D consistency for 3D GANs, several works \cite{trevithick2024you,chen2023mimic3d,xiang2023gram} proposed extensions to EG3D that allow for removing the super-resolution part after the NeRF rendering. This was mainly achieved by moving the super-resolution in front of the NeRF rendering. While this slows down the rendering, we observe considerably better 3D consistency in the rendered 3D heads. Nevertheless, since these methods use NeRF rendering, the generator model always outputs a 2D image of the implicit 3D model, making it difficult to use the results in explicit 3D environments. 

-subsection{3DGS GANs}
With the introduction of 3D Gaussian splatting (3DGS) \cite{kerbl3Dgaussians}, we obtained a fully explicit method which allows for fast rendering that is qualitatively similar to NeRFs. Ever since, many works have demonstrated very good rendering quality for synthesizing 3D human heads \cite{avatar1,avatar2,avatar3,avatar4}. Often, however, these methods rely on multi-view input data to overfit a single human head. This data is difficult to come by, as it either requires an expensive multi-camera setup \cite{kirschstein2023nersemble} or a video of a person remaining perfectly still. As this not required when training a 3D GAN, which learns 3D geometries from 2D data, there is a high demand for 3DGS based GANs. 
The first method to achieve 3DGS human head synthesis from a latent space is the GSGAN Decoder \cite{gsgandecoder}. It directly converts tri-plane features from a pre-trained EG3D model into Gaussian splatting scenes, enabling all advantages of GANs (i.e. synthesizing fakes, inverting or editing) to be employed in an explicit 3DGS setting. On the downside, however, the GSGAN Decoder adds additional computation, loses some of the detail and always requires a pre-trained EG3D. The second 3DGS GAN method, GGHead \cite{kirschstein2024gghead}, was proposed as the first method to train 3D human heads in a GAN training framework. Similarly to EG3D, GGHead first synthesizes 2D features using a StyleGAN2 backbone. However, instead of aligning the features in tri-plane and sampling the volume with a NeRF decoder, the output feature maps are used as UV-maps on a fixed template head mesh. Such an UV-map is constructed for each of the 3DGS attributes (color, position, scale, opacity, and rotation), where the position UV-map denotes an offset to a fixed position on the template mesh. As a result, GGHead's generator produces a 2D grid of Gaussian primitives that are used to model the surface of a head model. This method has demonstrated strong performance, producing competitive \textit{Fréchet Inception Distance (FID)} results while rendering with explicit 3D Gaussians. The third method to achieve Gaussian splatting GAN training is GSGAN \cite{hyun2024gsgan}. GSGAN follows a different approach to GGHead, as it does not model the surface on a head mesh, but directly predicts the Gaussian splatting scene using transformer models \cite{vaswani2017attention} that are specifically designed for point clouds \cite{guo2021point,zhao2021point,nichol2022point}. Similarly to a StyleGAN architecture, GSGAN produces coarse Gaussian primitives in earlier layers, while fine-detail Gaussians in subsequent layers. Finally, all Gaussian primitives from all layers are concatenated into a single 3DGS scene that can be rendered and forwarded to the discriminator. With this configuration, GSGAN finds a good balance between large Gaussians that model smooth regions like skin, and small Gaussians that model fine details like hair. To create a relationship between the layers, GSGAN builds a hierarchical data structure, where smaller Gaussian primitives are attached to larger ones from previous layers. As a result, GSGAN produces good quality 3D heads with competitive FID results. For our method, we will use GSGAN as a basis.

\section{Method}
\label{sec:method}

\begin{figure}
    \centering
    \includegraphics[width=1.0\linewidth]{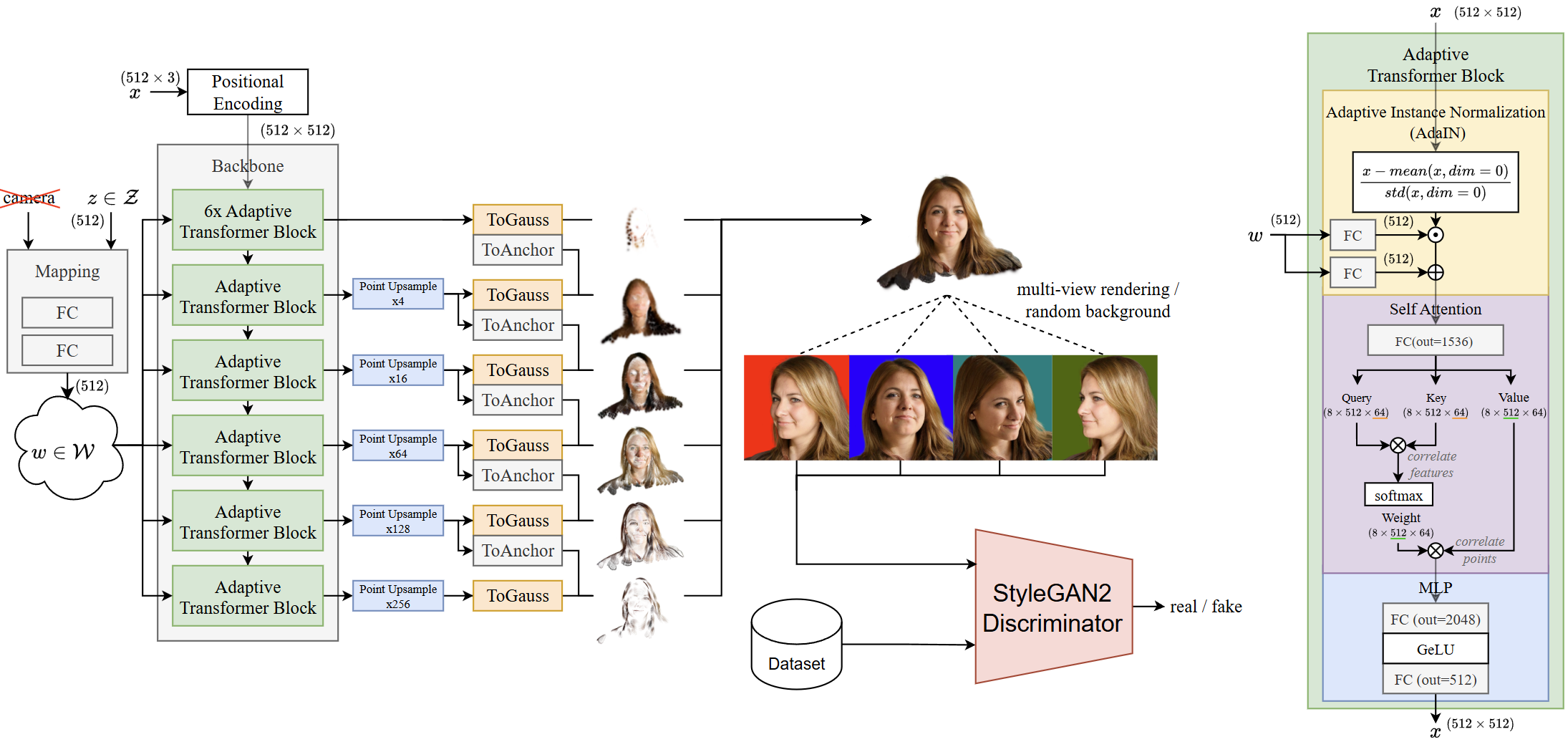}
    \caption{Overview of the proposed CGS-GAN framework. The generator is built on the generator of GSGAN \cite{hyun2024gsgan} with key modifications for improved 3D consistency and scalability.  In contrast to view-conditioned approaches, our method omits camera labels in the mapping network and instead stabilizes training through efficient multi-view rendering of the 3DGS head during each training step. Additionally, we render with random backgrounds to reduce hole artifacts.}
    \label{fig:architecture}
\end{figure}

In this section, we present the core components of CGS-GAN, our novel framework for high-fidelity, 3D-consistent synthesis of human heads using 3D Gaussian Splatting. Our method is designed to achieve stable training without relying on view-conditioning, while enabling efficient and scalable rendering. We first introduce our tailored generator architecture, which supports efficient high quality and high resolution rendering. We then  describe our multi-view regularization that enforces consistency across viewpoints during training. In addition, we propose a modified loss formulation adapted to our unconditional setting and introduce a background augmentation technique that mitigates hole artifacts and improves the geometric quality of the resulting 3D scenes.

\textbf{Generator Architecture}: Our generator architecture is designed to enable high-resolution, 3D-consistent head synthesis while maintaining efficiency and scalability. We use a  hierarchical generator structure similar to GSGAN \cite{hyun2024gsgan}. We first create a constant learnable feature point cloud of size $(512 \times 3)$ that is randomly initialized and forwarded to a positional encoding function that adds higher frequencies as in NeRF \cite {mildenhall2020nerf}, resulting in size $(512 \times 512)$. The encoded representation is then processed by a series of \textit{Adaptive Transformer} Blocks from GSGAN, where each block is composed of an \textit{Adaptive Instance Normalization Layer (AdaIN)}, a \textit{Self Attention Layer}, and a small \textit{Multi-Layer Perceptron (MLP)}. 
As shown in \autoref{fig:architecture} (right), the AdaIN layer normalizes the input features w.r.t.\ mean and standard deviation and applies a learned offset and scale, derived from a mapped latent vector $w\in \mathcal{W}$. Intuitively, this operation applies the \textit{style} of a latent vector to the learned feature point cloud representation. 
This idea originates from the well established StyleGAN \cite{Karras2019stylegan2} framework, known for its state-of-the-art rendering capabilities. The subsequent Self Attention layer enables the model to  establish meaningful correlation between the points. 
Finally, we apply a small MLP to create additional capacity \cite{vaswani2017attention}. From the resulting feature points, we decode a 3DGS scene by forwarding the points through fully connected layers with respective output channels for each attribute (position: 3, color: 3, scale: 3, rotation: 4 and opacity: 1) and apply $\tanh$ to clip the output to a reasonable range, while keeping differentiability. This is done for Gaussians that are rendered and Gaussians that are used as anchors or offsets for subsequent layers. To create the next layer of Gaussians, we again apply an adaptive transformer block to the feature point cloud of the previous layer and convert the result to a 3DGS scene that is offset and orientated by the anchors of the prior layer. This time, we apply point-upsampling to create four times more Gaussians than in previous layer. Notably, unlike GSGAN, we do not apply point upsampling within the backbone network. Instead, we increase the upsampling ratio for each subsequent layer. This is a very important aspect when it comes to scaling the generator for larger resolutions. Since, with our generator the number of points in the backbone is fixed, we can flexibly add additional layers to our generator, while scaling linearly in complexity. This is not the case for GSGAN, which performs point upsampling withing the backbone. By doing so, GSGAN cannot apply a transformer block to the second (or higher) layer, as the memory consumption would be too high. Instead, GSGAN applies graph convolution layers between neighboring Gaussians, which are very inefficient, as they require sorting distances between all points. As a result, our proposed generator requires substantially less GPU memory while being considerably faster. To scale our generator for higher resolutions, we add further transformer layers and double the point-upsampling factor. With this configuration, we end up with the following number of Gaussian primitives for the corresponding training resolutions: \{$256^2$: 109k, $512^2$: 240k, $1024^2$: 502k, $2048^2$: 1M\}.

\textbf{Multi-view Regularization}:
Training instability is one of the main problems when working with GANs. Usually, the generator cannot keep up with the performance of the discriminator, which causes the training to collapse. For 3D synthesis, this instability is amplified, since the generators job of constructing a 3D head is much more complex than the discriminators job of evaluating the realness of provided 2D images. 

To encounter this training instability, we introduce a multi-view regularization that renders the generated 3D heads from multiple views per training step and averages the resulting gradients before backpropagation. This brings two advantages. Firstly, as we average the gradient on a larger sample size, the direction of the gradient is less affected by stochastic variation, improving general convergence. And secondly, as we render the same scene from multiple views per training step, we collect gradients that do not push the weights towards one specific view per latent vector, but instead evenly balance the optimization across multiple views at the same time. This is especially important for 3DGS, given that the positional gradient is very sensitive to small changes \cite{gsgandecoder,kirschstein2024gghead,hyun2024gsgan}. Usually, when using NeRF-based GANs, like EG3D \cite{Chan2021}, such an operation would be very inefficient, as the rendering with NeRF is a huge bottleneck compared to the feature synthesis part. With the efficient 3D Gaussian splatting rasterizer, we have the opposite case, where the rendering at >200 FPS is considerably faster than the backbone synthesis. As a result, our multi-view regularization only creates a neglectable overhead to the overall training time. 

As the discriminator is equipped with a \textit{Minibatch-Standard-Deviation} layer that collects statistics across the batch dimension, we have to be careful not to alter the batch statistics in any way compared to the real batches. For instance, we cannot provide the discriminator with a batch showing four times the same person from different angles, given that no person appears even twice in the real training data. Instead, if we apply the multi-view regularizations four times, we create four new batches where each person is still only shown once per batch.

\textbf{Conditional Loss}: In order to learn 3D geometry from 2D data, the discriminator is provided with the annotated camera position to learn which camera position belongs to which appearance. Once the discriminator has learned the 3D concepts from real data, it provides meaningful feedback for the generator how to mimic the 3D geometry. However, the way in which this feedback is created can differ quite a lot. GSGAN for instance, applies a contrastive loss function that pulls together matching camera / image pairs, while pushing non-matching pairs apart. Although this approach has proven to work, we observe that it destabilizes the training after several million training iterations. This is caused by the large training set containing quite a lot of camera / image pairs that are very similar to each other. If we now sample a batch of images and cameras that have very similar camera positions, the discriminator will produce high correlation between those non-matching pairs. Although it seems to be not bad if the discriminator confuses two very similar camera labels, it will result in very high loss terms when applying the contrastive loss. Therefore, we argue that the contrastive loss formulation is not ideal for such an optimization problem and dataset. Instead, we use the conditional training mechanism from StyleGAN \cite{Karras2019stylegan2}. Specifically, we do not directly apply a loss on the camera parameters itself but embed the camera parameters into the image features of the discriminator, which are then used to predict whether the provided image is real or fake. This conditioning, without an explicit conditional loss function, is well established for 2D GANs as well as 3D GANs \cite{Chan2021,an2023panohead}.
While removing the contrastive loss, we keep both regularizations from GSGAN, which penalizes Gaussian primitives that move too far from the center and enforces the Gaussians to cluster by calculating a KNN distance between the anchors on the lowest level.

\begin{figure}[t]
    \centering
    \includegraphics[width=1.0\linewidth]{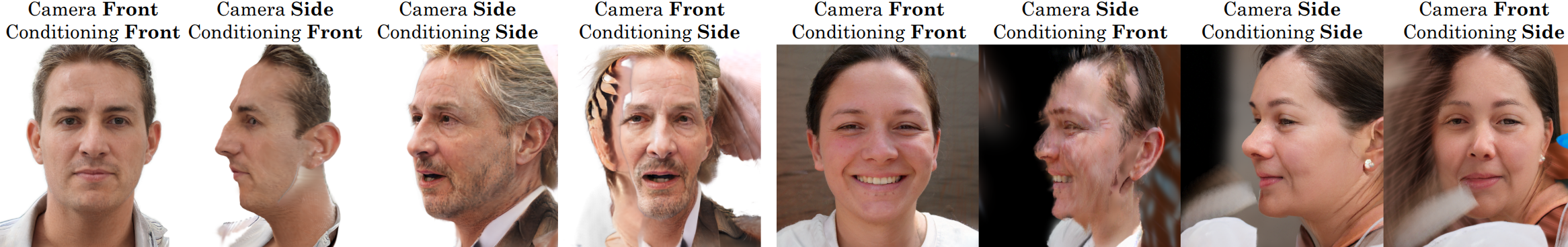}
    \caption{Visualizations how GGHead (left) and GSGAN (right) rely on  view-conditioning to achieve good quality renderings from any angle. As soon as the view-conditioning does not align with the camera pose, we observe worse quality. Additionally, we use a fixed latent vector for all four images, underlining that view-conditioning also changes the identity and expression.}
    \label{fig:conditioning_prob}
\end{figure}

\newpage
\textbf{3D Consistency}: While in theory, GSGAN \cite{hyun2024gsgan} and GGHead \cite{kirschstein2024gghead} produce 3D consistent human heads, they both make use of camera conditioning, where they input the camera viewpoint to the mapping network of the generator. Specifically, during training they forward in 50\% of the time the camera information to the mapping network of the generator. Although this was designed to give the generator the ability to deal with view-dependent biases in the dataset and therefore improve training convergence, we observe that the resulting generators heavily rely on this conditioning during inference as they produce poor quality renderings when conditioning the generator to a frontal view, while rendering from the side \autoref{fig:conditioning_prob}. Removing this view-conditioning typically destabilizes the training, even causing it to collapse in some cases. \textit{"The necessity for such tricks is one of the main disadvantages of 3D GANs and shows that we still need more research to find better 3D generative modeling paradigms"} \footnote{\url{https://github.com/tobias-kirschstein/gghead/issues/19}}. As our training is now much more stable with our improved generator architecture, the multi-view regularization and the removed contrastive loss, we are able to fully remove the camera conditioning from the training pipeline without harming training convergence. By doing so, the generator learns to synthesize a consistent 3D head that can be rendered from any viewpoint at high quality.

\textbf{Random Background}: Like GGHead, we remove the background in the training dataset using the background matting network, MODNet \cite{MODNet}. GGHead then replaces the background with a white background. As a result, we observe that the generator sometimes exploits the white background color and uses it to produce white regions, such as reflections, by creating holes in the surface. To remove this artifact, we replace the white background with a random colored background that changes for each new image sample. This way, the generator cannot rely on any color from the background, enforcing it to produce the correct color that is required for rendering the human head itself.

\section{Dataset}
\label{sec:dataset}

When training with in-the-wild images from FFHQ, we observe rendering artifacts that include large undefined floating objects in front of or next to the 3D head. Most often, these objects look like human hands, microphones, or additional human heads. They originate from the training data that includes several people with their hand covering the face, while drinking, smoking or holding a microphone.
As the generator is enforced to replicate the same data distribution provided in the training data, it has to mimic the occluded training examples as well. This is a problem for several reasons: Firstly, these occluders are frequent enough in the training data to force the generator to reproduce them, however, they are not frequent enough to give the generator enough information to render them realistically. As a result, they appear in many synthetic images and in poor quality.
Secondly, to avoid such artifacts during inference, often the truncation trick is applied. The truncation trick moves the randomly sampled latent vector towards the mean vector, which typically synthesizes faces with higher quality and fewer occluders but trade image quality with image diversity. Thirdly, rendering the occluders in addition to the 3D head uses capacity of the generator that could be spend on better face details. This especially applies for 3DGS GANs that use a fixed number of Gaussian primitives to construct the 3D scene. 
For all those reasons, we argue that the overall quality drastically benefits from removing these occluders from the training data. Even though this removes some variation, we argue that effectively more variation is kept, as we no longer have to apply a strong truncation to avoid unwanted artifacts. 
In total we remove 15k images which is about 20\% of the training data. To identify the images where occluders are present, we train a VGG-19 on a small training set curated with PicArrange \cite{jung2022picarrange}, an application that allows querying large datasets with CLIP prompts, such as: 'a person drinking'. As a result, we propose a new subset of FFHQ, \textit{FFHQ-Clean (FFHQC)} that only keeps training images without occluders.

\begin{figure}
    \centering
    \includegraphics[width=1.0\linewidth]{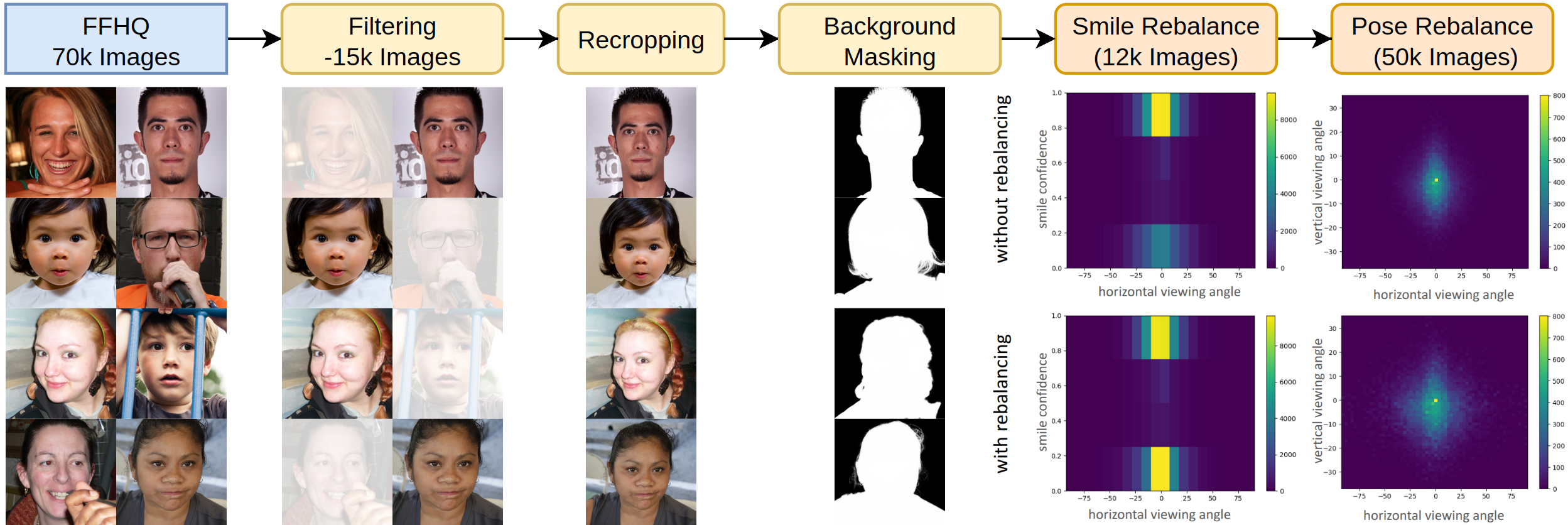}
    \caption{Our data processing pipeline: 1. remove images with occluders to enhance training, 2. recrop the entire head at $2048^2$ resolution, 3. apply background masking, 4. rebalance the smiling bias, and finally 5. rebalance the camera positions.}
    \label{fig:data_pipeline}
\end{figure}

We further observe that 3D consistency is slightly harmed when training our generator with the default distribution of FFHQ, since FFHQ has a strong bias for people smiling when looking into the camera. Since our model has to replicate this effect without being equipped with view-dependent components, it starts to align very thin Gaussian primitives in a way that a person smiles only when viewed from the front, harming the perceived 3D consistency. To avoid this, we rebalance the dataset by adding 12k images of persons viewed from the front who do not smile. Furthermore, we apply rebalancing for the camera view to improve the synthesis quality for side views, by duplicating 50k images from extreme poses. The full data processing pipeline in visualized in \autoref{fig:data_pipeline}.

\section{Experiments}
\label{sec:exp}

We compare our method with two existing Gaussian splatting based 3D GANs for human head synthesis, GGHead \cite{kirschstein2024gghead} and GSGAN \cite{hyun2024gsgan}. For GSGAN, we remove the background synthesis. We first train all methods with the default FFHQ dataset with removed background, and then apply our new dataset, FFHQC. A detailed training configuration can be found in the appendix. As we specifically set our focus on explicit 3D human head synthesis, we do not compare our results to implicit NeRF-based 3D aware-GANs. Such comparisons can be found in \cite{kirschstein2024gghead} and \cite{hyun2024gsgan}.

To evaluate and compare our results with prior methods, we use the Fréchet Inception Distance (FID) that compares the feature statistics, created with the pre-trained InceptionV3 network, of 50K rendered images to the feature statistics of the training data. Although the FID is a well established method for calculating the image quality, it does not account for 3D consistency, as it only generates a single image per 3D head. This highly favors those methods that render in very good quality for one specific view, but render poorly from any other view. In order to properly measure the quality of the synthesized 3D model, we propose a modified FID metric, $\textbf{FID}_{\textbf{3D}}$, that measures the quality of the 3D faces without exploiting view-conditioning. Specifically, we condition prior methods on the frontal view, given that it provides the best result for most views, but render from a randomly sampled view from the dataset. For our method without conditioning FID and $\text{FID}_{\text{3D}}$ are equal.

\begin{table}[b]
    \caption{FID (50kFull) results where the latent vector is conditioned on the current viewing angle (left table) and $\text{FID}_{\text{3D}}$ results where we use a fixed frontal conditioning for all images (right table).}
    \parbox{.5\linewidth}{
        \centering
        \begin{tabular}{|l|r|rrr|}
            \hline
            \textbf{FID} & FFHQ & \multicolumn{3}{|c|}{FFHQ Clean} \\
            & 512 & 512 & 1024 &  2048 \\
            \hline
            \hline

            GSGAN  & 5.02 & 5.17 & / & / \\
            GGHead & \textbf{4.34} & 5.37 & 9.91 & / \\ 
            Ours   & 4.94 & \textbf{4.53} &  \textbf{5.25} & 7.8 \\
            \hline
        \end{tabular}
    }
    \parbox{.5\linewidth}{
        \centering
        \begin{tabular}{|l|r|rrr|}
            \hline
            $\textbf{FID}_{\textbf{3D}}$ & FFHQ & \multicolumn{3}{|c|}{FFHQ Clean} \\
            & 512 & 512 & 1024 &  2048 \\
            \hline
            \hline
            GSGAN  & 10.50 & 7.68 & / & / \\
            GGHead & 7.90  & 7.78 & 14.27 & / \\
            Ours   & \textbf{4.94}  & \textbf{4.53} & \textbf{5.25} & 7.8 \\
            \hline
        \end{tabular}
    }

    \label{tab:fid}
\end{table}

\textbf{Quantitative Results:} In \autoref{tab:fid}, we compare the FID and $\text{FID}_{\text{3D}}$ results to prior Gaussian Splatting based 3D GAN methods. For the standard FID, that allows for view-dependent rendering, our model achieves comparable scores to GGHead and GSGAN, with GGHead achieving the best score of 4.34 for FFHQ. For our proposed dataset FFHQC, however, our model performs better than both prior methods. Here, GGHead performs worst, because of the template mesh that is optimized for frontal heads. As FFHQC includes full heads and also more views from the side, the provided template mesh might not be suitable anymore. 

For the  $\text{FID}_{\text{3D}}$ results, we observe significantly worse quality for GGHead and GSGAN. This is because, the side views now have poor quality, as we restrict the models to produce consistent scenes conditioned on the frontal view. For resolutions of $1024^2$ and $2048^2$ our model produces low values, indicating good scaling capabilities and robust training convergence.


\begin{figure}[t]
    \centering
    \includegraphics[width=0.99\linewidth]{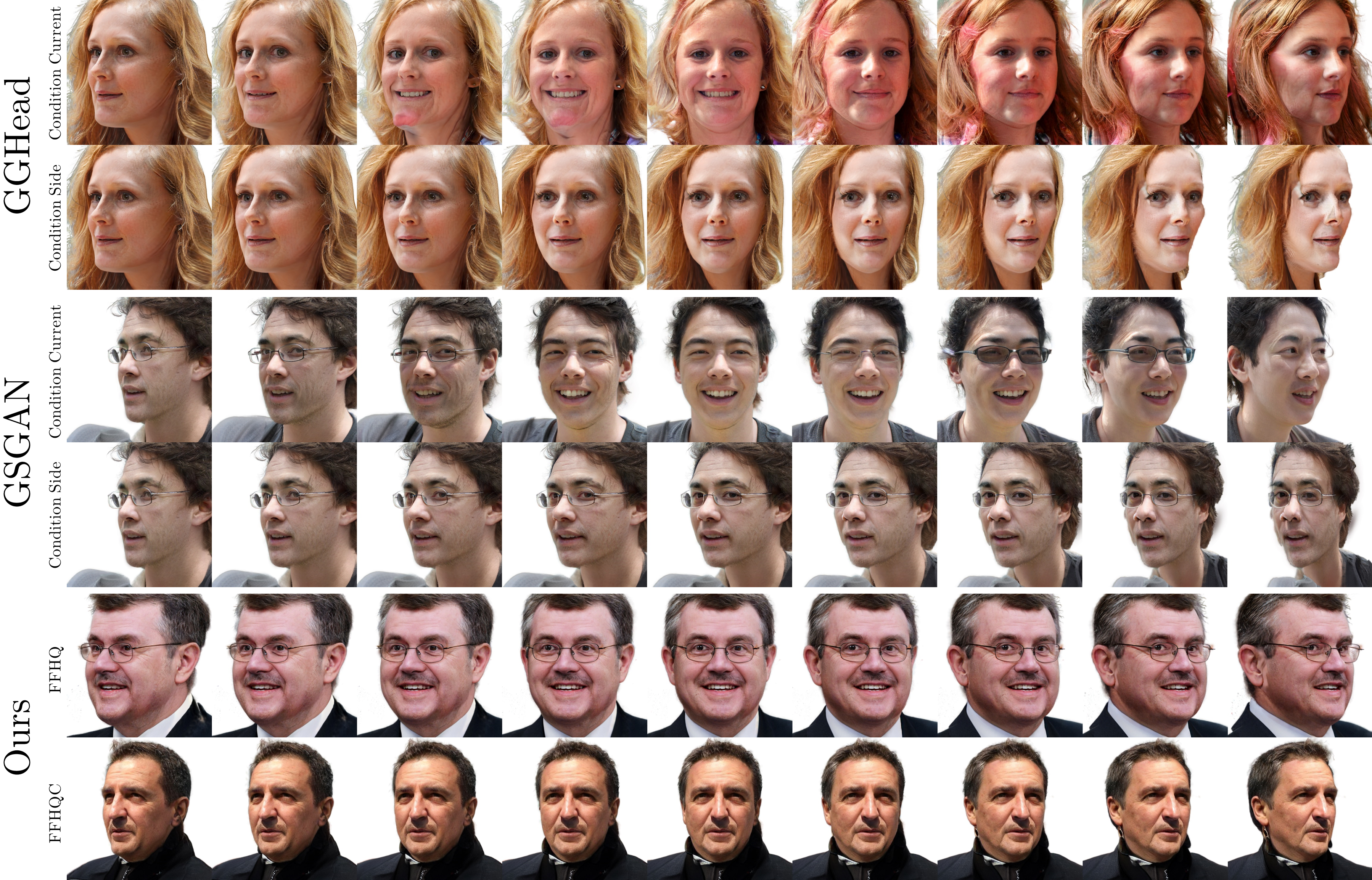}
    \caption{A comparison between GGHead, GSGAN and our proposed method. For prior methods we apply conditioning on the current view (first rows), resulting in good but inconsistent identities, and apply conditioning on a left side view (bottom rows), resulting in a consistent scene with poor quality for novel views. In contrast, our method creates high quality renderings for the whole rotation for FFHQ and FFHQC (last row).}
    \label{fig:compare_180_renderings_main}
\end{figure}

\textbf{Qualitative Results:} \autoref{fig:compare_180_renderings_main} shows example renderings of our method compared to prior methods. It illustrates how the view-conditioning helps synthesizing good renderings for steep angles however at the cost of change in person identity. For GGHead and GSGAN, we observe changes in hair color, expression and glasses when moving the camera without altering the latent vector. Alternatively, if we condition a fixed latent vector, in this case the left most image highlighted in red, we observe good quality for that view, but worse quality the further we rotate. Both problems are no longer present with our proposed method that removes the view-conditioning. Our method produces high quality renderings from any viewpoint, producing a single fixed 3DGS scene. Results for $1024^2$ training are shown in \autoref{fig:teaser}. We provide more qualitative results in the appendix.

\textbf{Performance and Memory:} In \autoref{tab:performance}, we compare the synthesis speed, GPU memory usage and training speed among the 3DGS GAN methods. Instead of measuring the rendering speed of the resulting 3DGS scene, which is very fast for all methods (> 200 FPS) leveraging the efficient 3DSG rasterizer, we measure the time for synthesizing novel faces. I.e., the time it takes between forwarding a random latent vector and rendering the resulting 3D scene. With GGHead being the overall fastest with 168 FPS, our method achieves comparable results with 153 FPS. GSGAN compares as the slowest, which is due to expensive graph convolution layers. Similar results are observed in training GPU memory, where our method and GGHead require more than half of the memory of GSGAN. For higher resolutions, our model still requires comparably little GPU memory. 

Although one training step of our method is slower than for GGHead, our method converges much faster. While GGHead trains for 25M iterations, taking about 4.5 days, our method already converges in less than 10M steps only taking less than 3 days. 

\begin{table}
    \centering
    \caption{Performance comparison between 3DGS GANs. Synthesis speed is measured on a Nvidia 4090. Training speed and GPU memory is measured using 4 Nvidia H100 with a batch size of 8.}

    \begin{tabular}{l|rrr}
        & Synthesis Speed & Training GPU Memory & Training Speed\\
        \hline
        \hline
        GSGAN@512       & 35.0 FPS & 37 GB & 43 sec / kimg\\
        \hline
        GGHead@512      & 167.7 FPS & 15 GB & 16 sec / kimg\\
        GGHead@1024      & 151.7 FPS & 36 GB & 27 sec / kimg\\
        \hline
        Ours@512     & 152.5 FPS & 15 GB & 25 sec / kimg \\
        Ours@1024    & 110.7 FPS &  19 GB & 64 sec / kimg\\
        Ours@2048    & 61.3 FPS  & 38 GB & 157 sec / kimg\\

    \end{tabular}
    \label{tab:performance}
\end{table}

\textbf{Multi-view Regularization}: Training our model without the multi-view regularization is significantly less stable, causing it to collapse after about 8M training steps. 
Further, as we do not generate new 3D scenes during multi-view regularization, the only computational overhead is the efficient 3DGS rendering and additional forward passes through the 2D discriminator. While generating 4x more images during training, the overall duration is only extended by 20\%. We found rendering 4 views per training step a good setting, as we do not improve using 8 and perform worse using 2.

\begin{figure}[h!]
    \centering
    \includegraphics[width=1\linewidth]{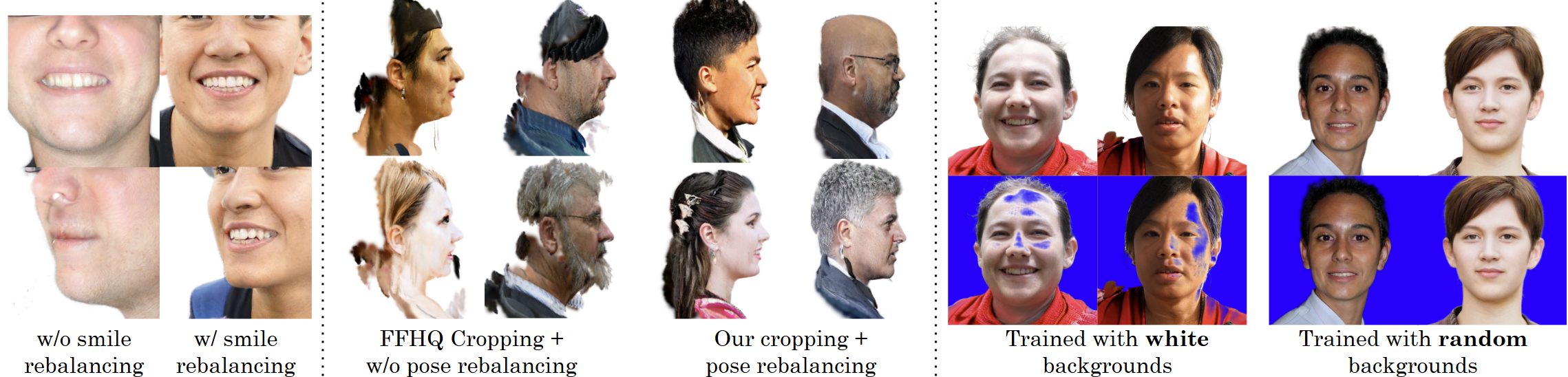}
    \caption{Left: Renderings w/ and w/o smiling bias. Center: Improved side views by training with entire heads and rebalancing the camera pose. Right: Training w/ and w/o random backgrounds.}
    \label{fig:ablation}
\end{figure}

\textbf{Random Background Augmentation}: In \autoref{fig:ablation} (right), we demonstrate the effectiveness of our random background augmentation. Without random backgrounds, the model overfits to the white background causing it to create holes instead of creating white Gaussian primitives. As our method can no longer rely on a single background color, it produces every color, required for rendering the 3D head, with a Gaussian primitive.

\textbf{Dataset Comparison:} Rendering images with our proposed dataset, FFHQC, shows to improve the overall quality. Not only are we able to synthesize the full head (\autoref{fig:compare_180_renderings} bottom) at high resolutions, we also observe better quality when rendering steep angles as shown in \autoref{fig:ablation} (center). This is caused by our camera rebalancing strategy, enforcing steeper angles to be shown more frequently. Furthermore, as we filter 15k occluded face images, we observe almost exclusively good quality 3D human heads, without rendering artifacts. Only in rare cases, we still observe some floating objects next to the face. This is likely due to the background matting network, which sometimes confuses background and foreground. Next to removed occluders, we observe that the smiling rebalancing resolved a problem where some rendered faces were smiling when shown from a frontal view, while looking neutral from a side view as shown in \autoref{fig:ablation} (left).


\section{Conclusion \& Outlook}
\label{sec:conclusion}

We propose a novel Gaussian splatting GAN framework, \textbf{CGS-GAN}, that is capable of efficiently synthesizing 3D consistent human heads at high-quality and high resolutions of up to $2048^2$. While prior work suffers from either rendering inconsistent persons or producing 3D heads that show poor quality for some viewpoints, our method achieves very good quality from any viewing angle, while producing a fixed 3DGS scene that can be ported into explicit 3D environments. This allows for applications such as inverting a 3D head from a 2D photograph and directly importing it into a game engine. This potentially saves time and resources, as no high-end 3D capturing studio is required. A straight forward extension to this work could be the inclusion of further training samples showing the back of the head, like in \cite{an2023panohead}, allowing for full 360° head synthesis. Furthermore, even though our method already trains very fast, requiring less iterations while possessing an efficient forward pass, we could apply optimized attention implementations to reduce the training time even further, making development of novel 3D GAN methods easier and more accessible. Finally, combining our method with 3D morphable models could turn our static human heads into animatable talking 3D head avatars.



\begin{ack}
This work has partly been funded by the German Research Foundation (project 3DIL, grant
no.\ 502864329), the European Commission (Horizon Europe project Luminous, grant no.\ 101135724) as well as the Fraunhofer Society in the Max Planck-Fraunhofer
collaboration project NeuroHum.
\end{ack}


{
    \small
    \bibliographystyle{abbrv}
    \bibliography{main}
}


\newpage
\appendix
\newpage
\section{Technical Appendices and Supplementary Material}
In the following sections, we give a more training details of our GAN framework \ref{sec:details}, provide further information about the components of the generator \ref{sec:arch}, discuss the limitations and broader impacts of this work \ref{sec:limitations}, make further comparisons to prior NeRF-based approaches \ref{sec:further_comparisons}, render further results \ref{sec:renderings}, and show some failure cases with possible solutions for future work \ref{sec:failure_cases}. Additionally, we give more details about the contrastive loss function used in GSGAN and show how it destabilizes the training \ref{sec:cont_loss}, give more details about the dataset filtering process \ref{sec:dataset_filtering}, and, finally, show that the resulting scenes can be used in explicit environments such as Unity \ref{sec:explicit}. More interactive results can be viewed on the project page that we include in the supplementary materials.

\subsection{Training Details}
\label{sec:details}
Given that GAN training can be very difficult to set up, a lot of works, including this one, base their implementation on an already well functioning GAN frameworks. While there are often significant changes for the generator architecture, we observer very similar hyperparameter settings, adversarial loss functions and regularization. Chronologically, there has been the \textit{Progressive Growing of GANs} \cite{karras2017progressive}, then \textit{StyleGAN} \cite{karras2019style} and \textit{StyleGAN2} \cite{Karras2019stylegan2}, and specifically for 3D rendering \textit{EG3D} \cite{Chan2021}. Those works all build upon the prior one, thus inheriting most training setting. The same goes for GSGAN \cite{hyun2024gsgan}, which we use as a starting point for our work, that applies most settings from EG3D.

As a result, our training settings are closely orientated to all prior methods. Specifically, this includes an Adam optimizer \cite{Kingma2014AdamAM} with a learning rate of 0.0025 for the generator and 0.002 for the discriminator and a R1 regularization of 1.0, which penalizes the discriminator if it creates too large gradients for real images. At the beginning of the training, we blur the real and fake images for a period of 200k training iterations, to enable smooth training convergence. We set the batch size to 32, which is collated of 4 batches of size 8 for each of the GPUs. During training, we track the FID with 20k fake images to identify the best checkpoint. Further, we average the generator weights in an \textit{Exponential Moving Average} EMA generator, which produces better results than the training generator. For our 2D discriminator, we use the StyleGAN2 discriminator from \cite{Karras2019stylegan2}.

Different to GSGAN, which represents the 3DGS scene in a 3D cube with side length of 2 (-1 to 1), we go back to the EG3D representation, using a cube with side legth of 1 (-0.5 to 0.5). GSGAN, also modifies the field of view of the rendering camera from 12° to 22°, given that it uses a much larger scene. We, however, also revert this setting to the original 12° used in EG3D. 

\begin{figure}[b]
    \centering
    \includegraphics[width=0.8\linewidth]{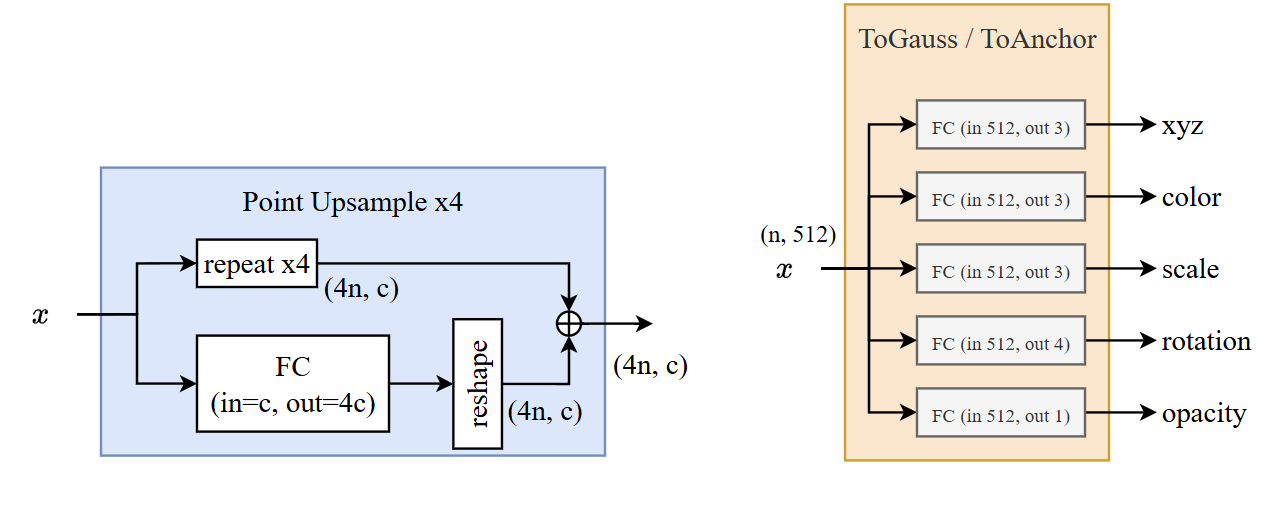}
    \caption{Overview of the point upsampling layer (left) and the toGauss / toAnchor layer (right) used in our generator.}
    \label{fig:gan_info}
\end{figure}

\subsection{GAN Architecture}
\label{sec:arch}
In \autoref{fig:gan_info}, we provide an overview of the point upsampling layer and the toGauss / toAnchor layer that we use in our generator architecture. The point upsampling is constructed of a repeat branch that simply duplicates the points four times and a learnable branch that produces four times more features and then reshapes the features into the point dimension. Intuitively, each point is the offspring to four new points with slightly changed features each. The toGauss / toAnchor is simply constructed of fully connected layers that map each high dimensional 512D point into a 3DGS feature. Additionally we also apply a weak tanh activation between -20 and 20. Otherwise, features such as rotations infinitely rise to values of 1000. For the rendered scene, this makes no difference since the magnitude of the quaternion is neglected, however, for achieving stable gradients it can make a big difference.

\subsection{Limitations \& Broader Impacts}
\label{sec:limitations}
Although our proposed model is capable of synthesizing high-fidelity renderings showing the full head, we still cannot render the back of the head, as such images are not provided in FFHQ. Including additional images showing the back of the head like in \cite{an2023panohead}, would likely achieve even better results. Further, even after removing 15k images showing occluders, we still observe some artifacts being present in some images. This needs further attention, either by filtering more images or improving the background matting network that sometimes fails to differentiate hats from backgrounds or masks multiple persons in an image. Additionally, our high resolution methods are still slow to compute, making further development in this field not accessible for many smaller research groups with fewer computational capacity. Therefore, we are aiming to apply more efficient CUDA implementations to reduce the training time even further. Further, we provide our training checkpoints for all resolutions, together with the optimizer momentum values, allowing to finetune our models without the necessity of re-training the full pipeline from scratch.

Building automatic tools for synthesizing realistic human heads is both a scientific advancement with many applications in the industry while also a dangerous tool that can be misused for producing fraud information. This is an ongoing and complex problem that needs more attention. Furthermore, FFHQ has a strong ethnic bias, which is very problematic when applying such methods in production. Finally, using such generative models to produce art, can be problematic as the training data might be used against the consent of the respective artist.

\subsection{Further Comparisons}
\label{sec:further_comparisons}
In \autoref{tab:more_comparisons} we provide further comparisons to NeRF-based approaches, such as EG3D and Mimic3D, while also another Gaussian splatting based method, \textit{Gaussian Shell Maps} (GSM), which was originally designed for full body 3D human avatars. We observe that EG3D produces the overall best results. Nevertheless, all except our method takes advantage of the view conditioning method. Further, EG3D also applies 2D super-resolution layers after NeRF rendering, making it even less 3D consistent.

\begin{table}[h]
    \caption{Further comparisons to EG3D, \textit{Gaussian Shell Maps} (GSM) and Mimic3D. This was calculated with FFHQ at 512 resolution with masked backgrounds. Our method is the only one that does not apply view-conditioning.}
    \centering
    \begin{tabular}{|l|r|}
        \hline
         Method & FID \\
         \hline
         EG3D \cite{Chan2021} & 3.28 \\
         Mimic3D \cite{chen2023mimic3d} & 4.27 \\
         \hline
         GSM \cite{gsm} & 28.19 \\
         Ours & 4.94  \\
         \hline
    \end{tabular}
    \label{tab:more_comparisons}
\end{table}

\subsection{Rendering Results}
\label{sec:renderings}
\begin{figure}
    \centering
    \includegraphics[width=1\linewidth]{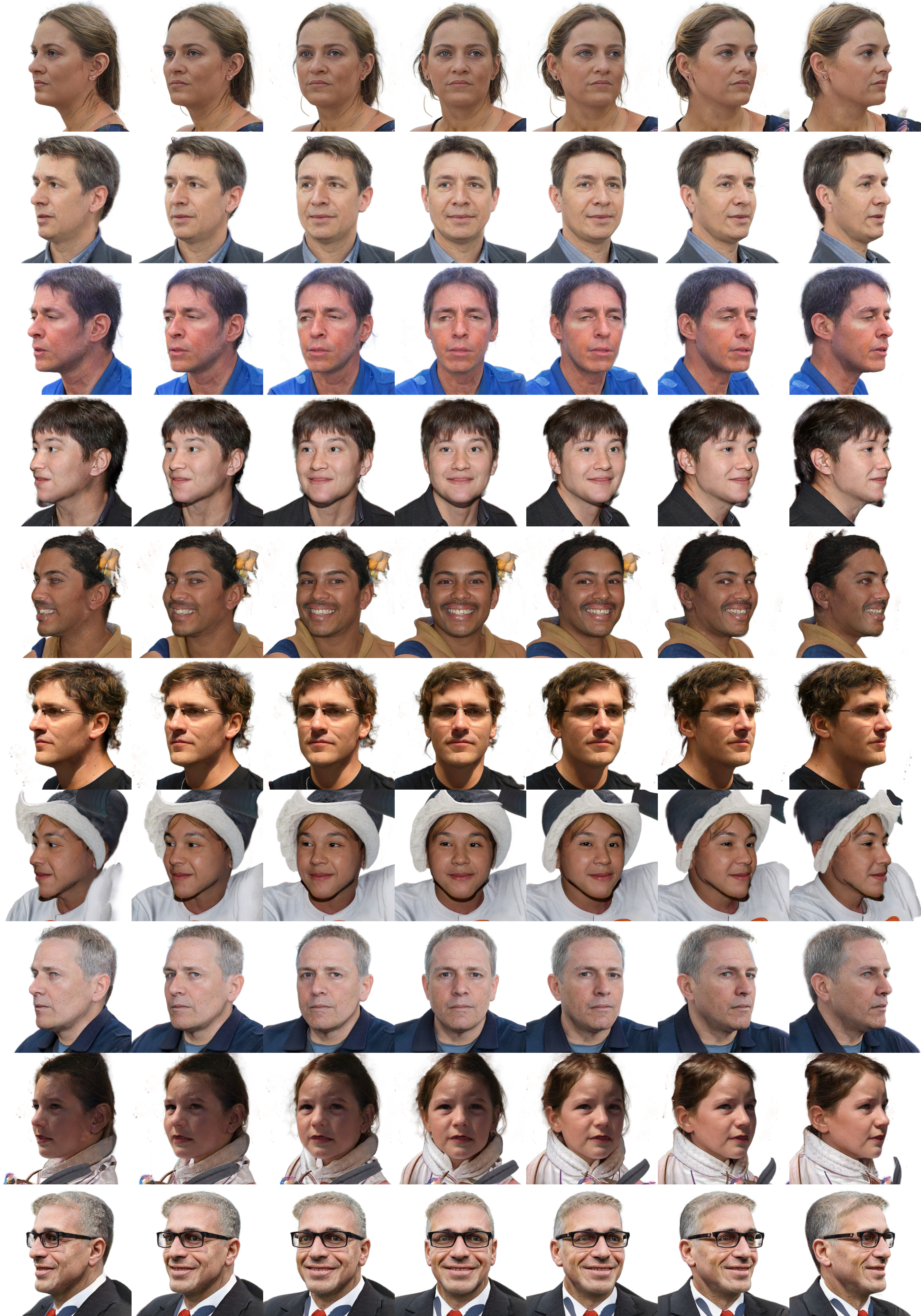}
    \caption{180° example renderings of our method trained on FFHQC at 512 resolution ($\psi = 0.8$).}
    \label{fig:compare_180_renderings}
\end{figure}

\begin{figure}
    \centering
    \includegraphics[width=1\linewidth]{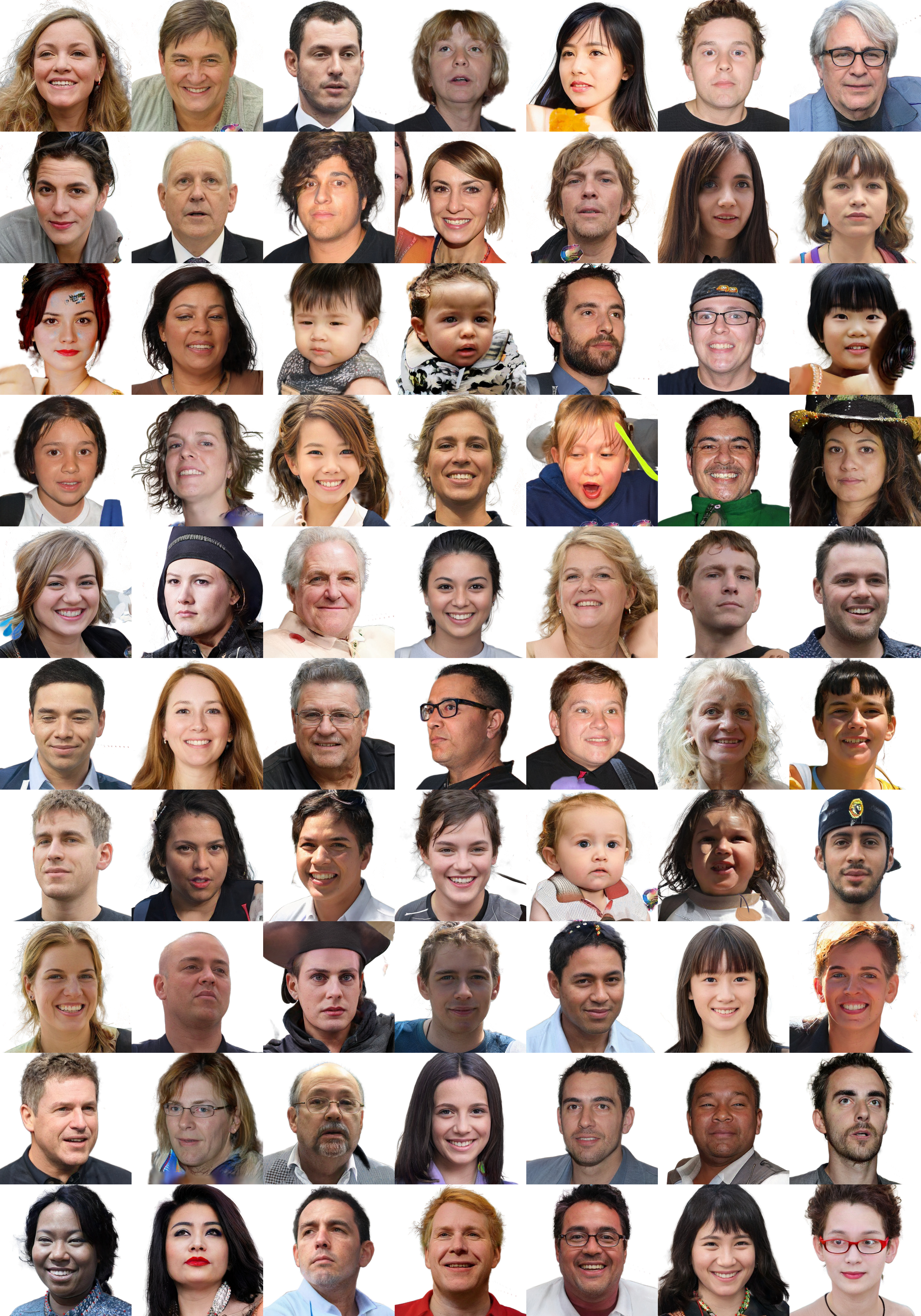}
    \caption{Example renderings of our method trained on FFHQC at 512 resolution ($\psi = 0.8$).}
    \label{fig:examples_random}
\end{figure}

\begin{figure}
    \centering
    \includegraphics[width=1\linewidth]{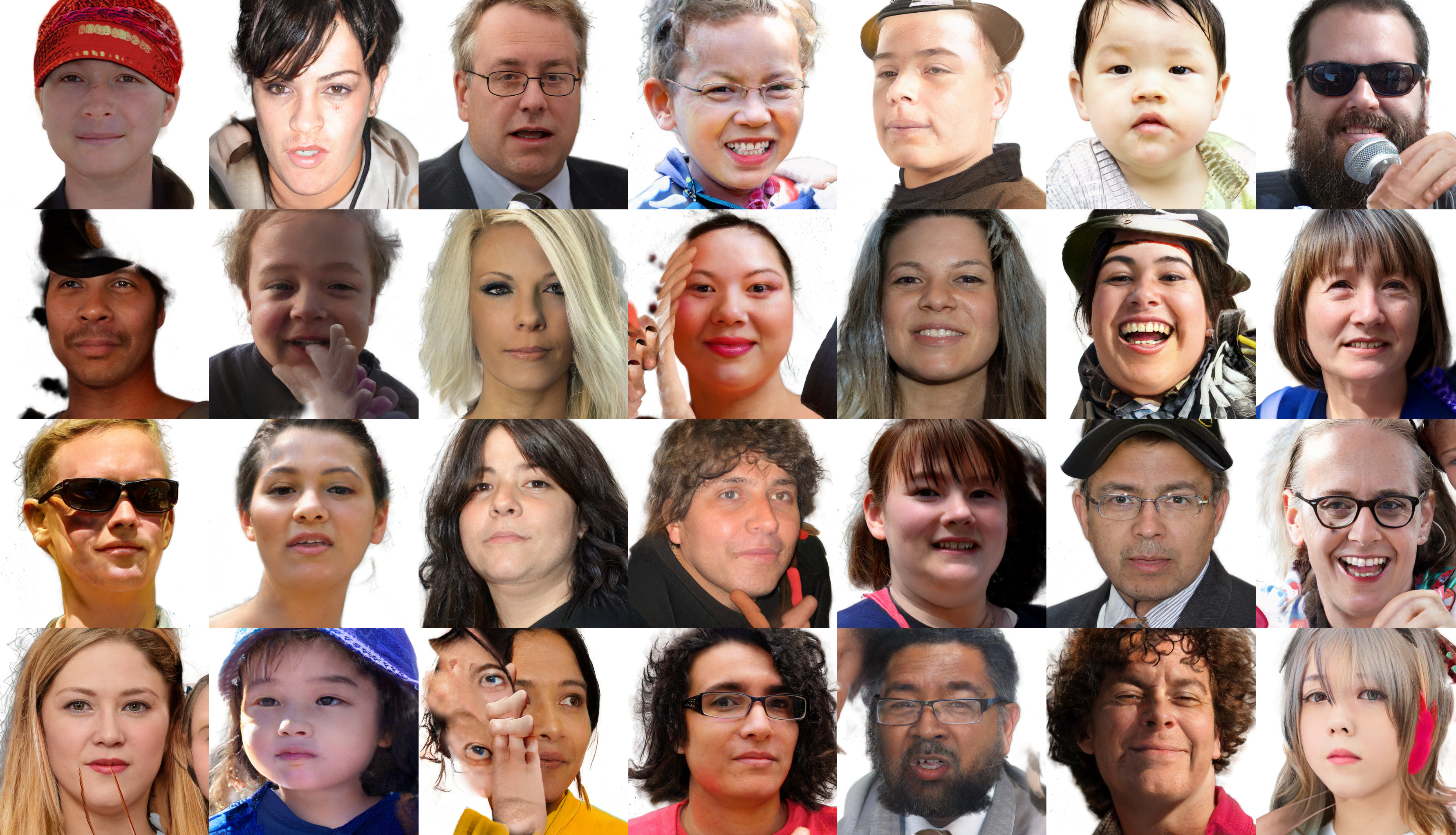}
    \caption{Rendering examples of our model with \textbf{FFHQ} ($\psi = 1.0$).}
    \label{fig:ffhq}
\end{figure}

\begin{figure}
    \centering
    \includegraphics[width=1\linewidth]{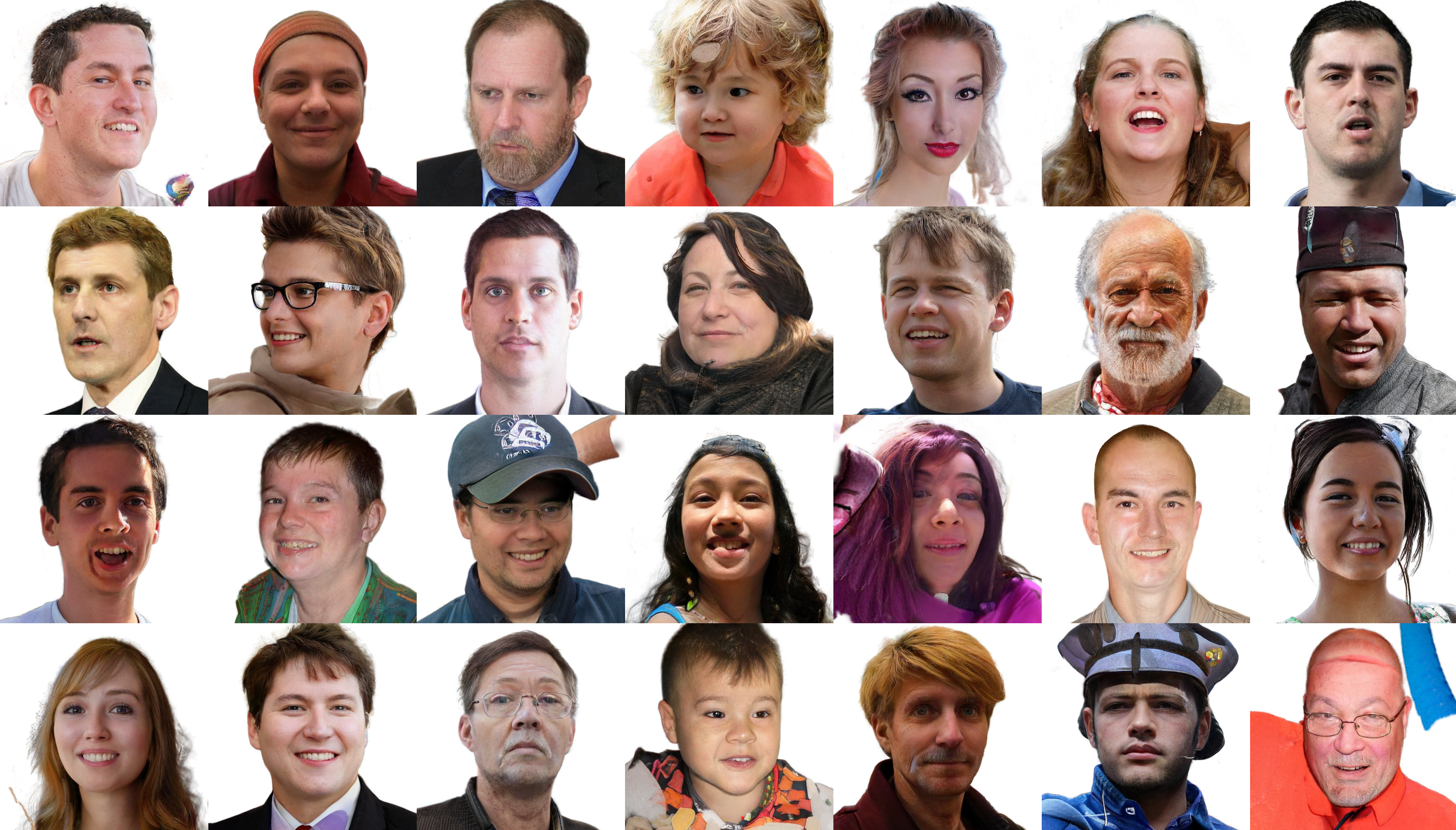}
    \caption{Rendering examples of our model with \textbf{FFHQC} showing the full head with fewer occluders ($\psi = 1.0$).}
    \label{fig:ffhqc}
\end{figure}

In \autoref{fig:compare_180_renderings} and \autoref{fig:examples_random}, we demonstrate uncurated renderings of our proposed method, trained on FFHQC at 512 resolution. Apart from some outliers, we observe very high quality with few occluders being present. Further, in \autoref{fig:ffhq} and \autoref{fig:ffhqc} we demonstrate the difference between the original FFHQ dataset and our adapted FFHQC dataset. We observe fewer occluders and render the full head instead of clipping just above the hair line.

\newpage
\subsection{Failure Cases}
\label{sec:failure_cases}
\begin{figure}[h]
    \centering
    \includegraphics[width=0.8\linewidth]{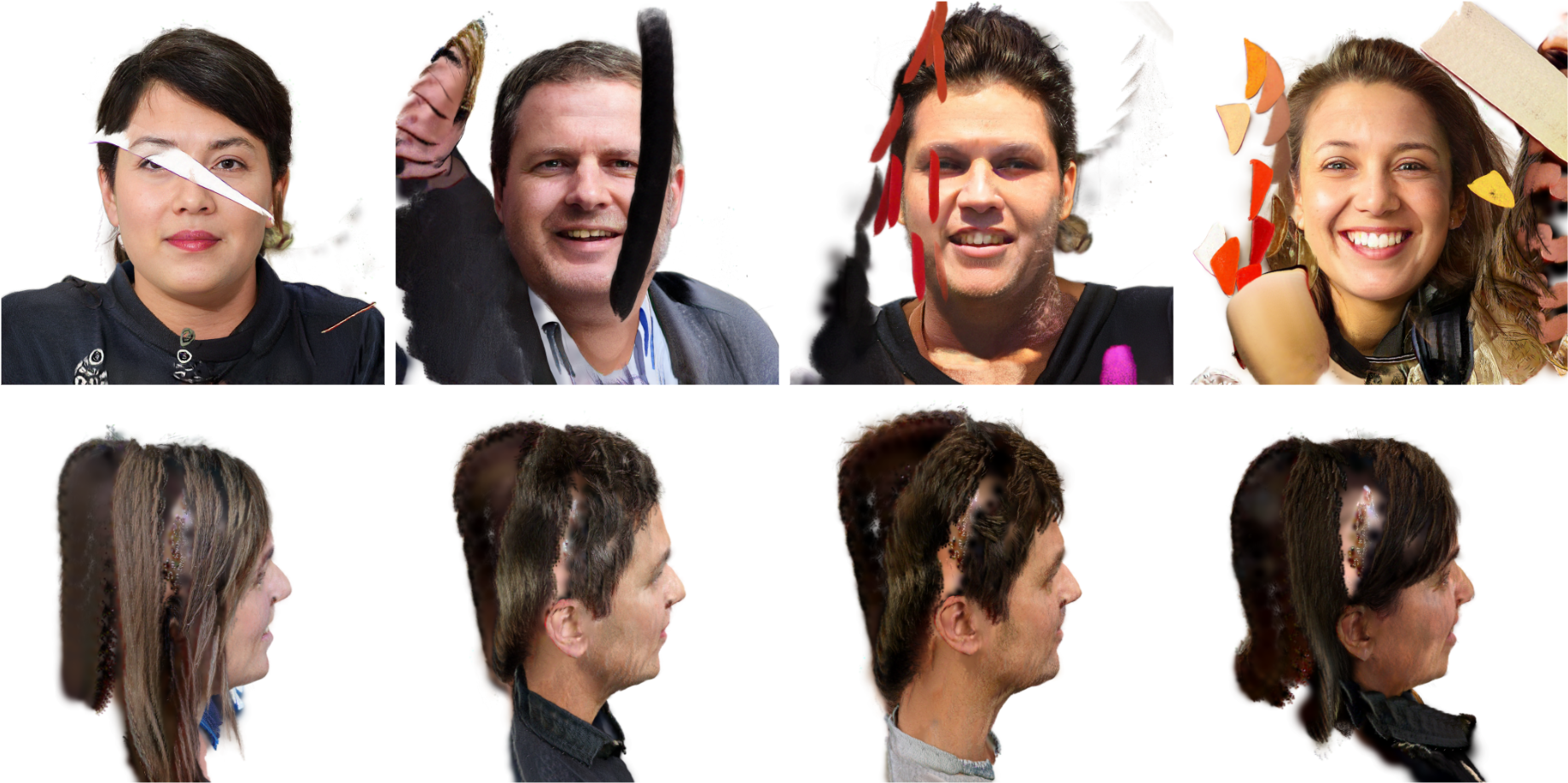}
    \caption{Top row shows some rare cases where the face is still obscured by floating objects. This is likely fixed by using a better masking network. And the bottom row shows holes that become visible when turning >90°. Such viewing angles are never shown during training.}
    \label{fig:bad_renderings}
\end{figure}

\begin{figure}[h]
    \centering
    \includegraphics[width=1\linewidth]{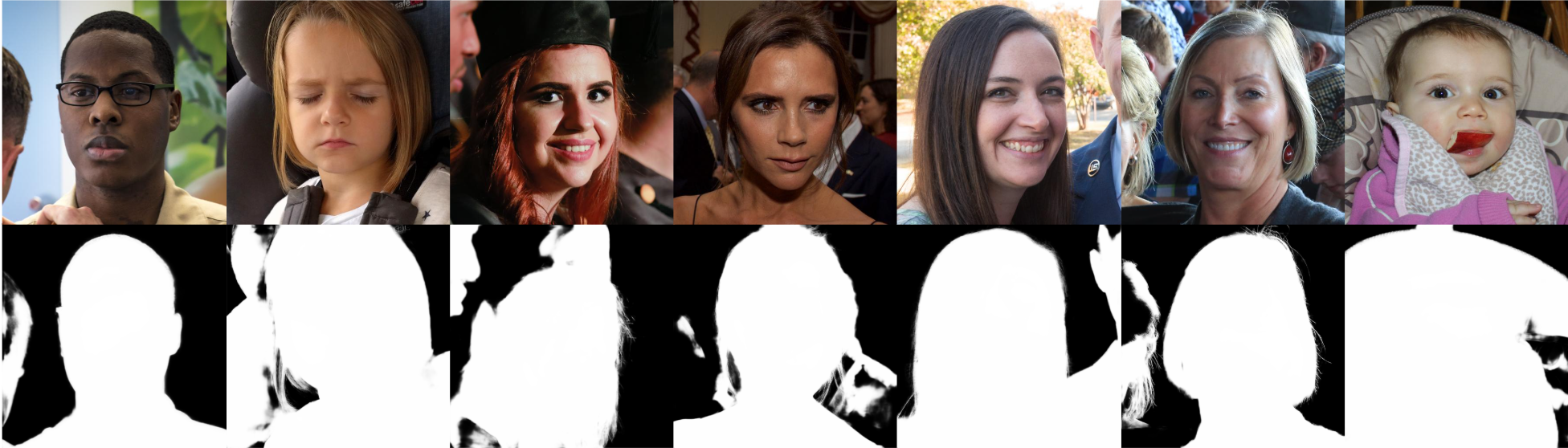}
    \caption{Real training examples with respective masks, where the masking network masks multiple persons, or where the background is confused with a hat.}
    \label{fig:bad_masking}
\end{figure}

While the vast majority of rendered images look very good, in some cases we still observe occluders to be present in the rendered output images. This is shown in the top row of \autoref{fig:bad_renderings}. We believe that this is due to the masking network that sometimes masks multiple persons or confuses the background with a hat as shown in \autoref{fig:bad_masking}. Additionally, we notice that some heads have a hole in the hair region when rotating them very far to the side. Such effect can likely be avoided by using more training samples with steeper viewing angles.

\subsection{Contrastive Loss}
\label{sec:cont_loss}
In the method section (Section 3) of our main paper, we argue that the contrastive loss function of GSGAN \cite{hyun2024gsgan} is not ideal for 3D GAN training with FFHQ. This is because a lot of camera view are very similar in the dataset, and the network is highly penalized if it confuses two almost identical camera views. Specifically, the contrastive loss if formalized as \cite{hyun2024gsgan}:

\begin{equation}
    \mathcal{L}_\text{pose} = -\log\left( \frac{\exp(\text{sim}(p_I, p^+_\theta ) / \tau}{\sum^B_{b=1}\exp(\text{sim}(p_I, p^+_\theta ) / \tau}\right).
\end{equation}

\begin{figure}[h]
    \centering
    \includegraphics[width=0.6\linewidth]{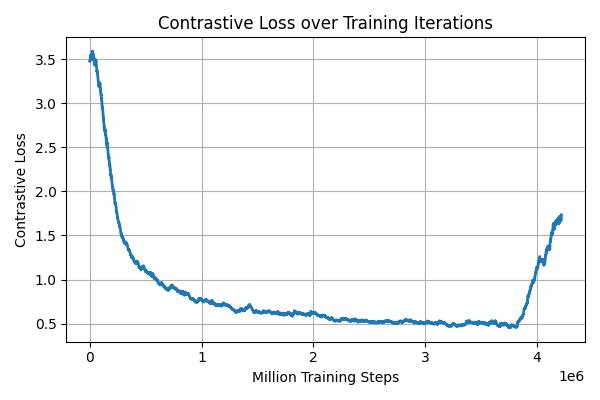}
    \caption{Contrastive loss tracked over the course of the training. After 3.8 million steps, the loss increases again, causing the training to collapse.}
    \label{fig:contrastive_loss}
\end{figure}

Here, \textit{sim} denotes the cosine similarity, $p_I$ an embedding of the image, $p^+_\theta$ the camera embedding of the matching camera,  $p^b_\theta$ the camera embedding of all non matching cameras in batch $B$, and $\tau$ a temperature scaling parameter. If we now have a $p^+_\theta$ that is very similar to another camera embedding $p^b_\theta$ within the training batch, we receive a very high loss that harm the training convergence. This can be seen in \autoref{fig:contrastive_loss}, where the contrastive loss decreases for the first few million training steps and then suddenly increases again after about 3.8 million steps. At this point the training completely collapses and does not recover again.

\subsection{Dataset Filtering}
\label{sec:dataset_filtering}

\begin{figure}[h]
    \centering
    \includegraphics[width=1.0\linewidth]{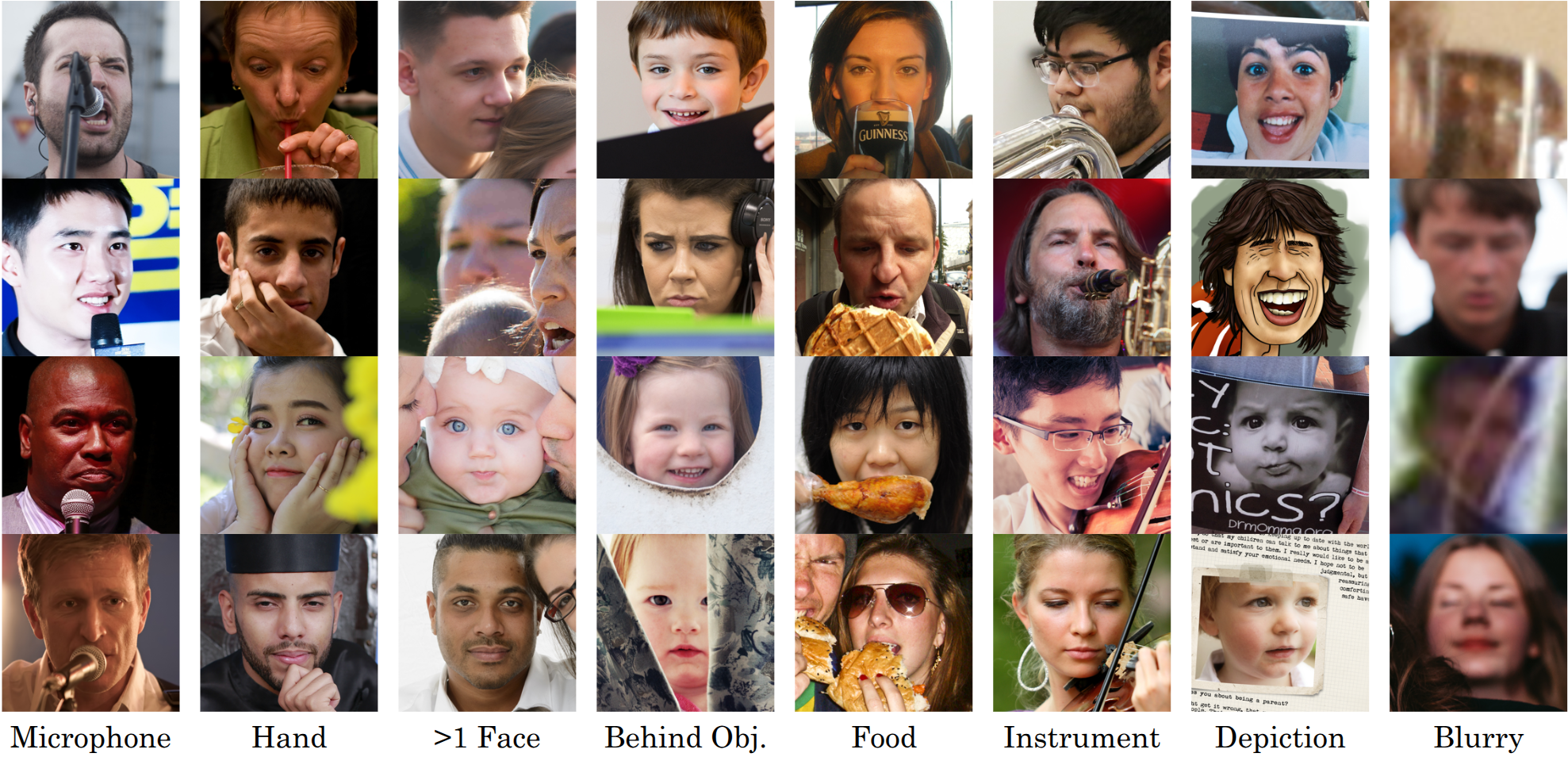}
    \caption{An overview of the main categories for images that we identify as bad training data. Categories like 'Hand', '>1 Face' and 'Microphone' were the most frequent ones.}
    \label{fig:categories_filter}
\end{figure}

In \autoref{fig:categories_filter}, we give an overview of the categories that we applied to identify bad training images. 
While mainly focusing on images where the face is obscured by an object, we also found images that generally harm the training convergence. Examples for such images are shown in the 'depiction' class and the 'blurry' class in \autoref{fig:categories_filter}.
Consequently, we removed those images along with the other images, with occluders. 

\begin{figure}
    \centering
    \includegraphics[width=1.0\linewidth]{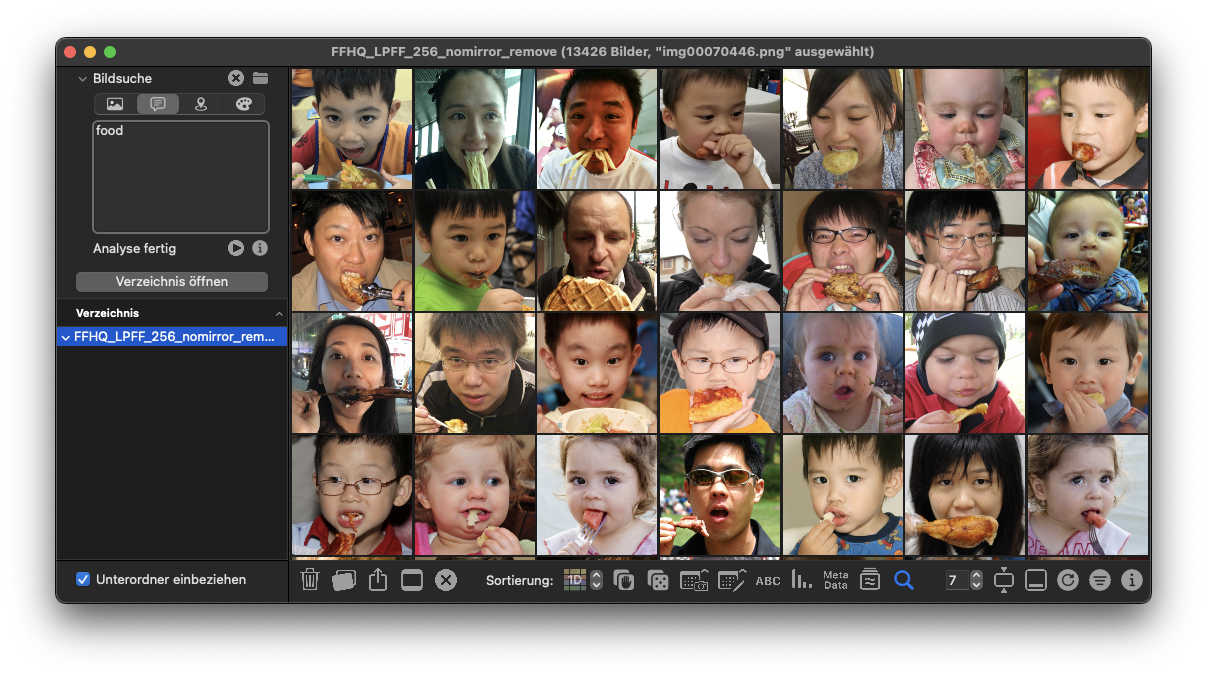}
    \caption{Screenshot of PicArrange \cite{jung2022picarrange}, which was used to identify bad bad training images, by writing text prompts like "food" in the upper left corner. PicArrange then returns the images with the highest CLIP correlation to the respective prompt.}
    \label{fig:picarrange}
\end{figure}

To identify, which images should be removed we use PicArrange \cite{jung2022picarrange}, which is an application that allows for querying large datasets with text prompts. A screenshot of the application is shown in \autoref{fig:picarrange}, where we query the dataset with the prompt 'food' and receive all matching images.

\subsection{Explicit 3D Environments}
\label{sec:explicit}
\begin{figure}[h!]
    \centering
    \includegraphics[width=1\linewidth]{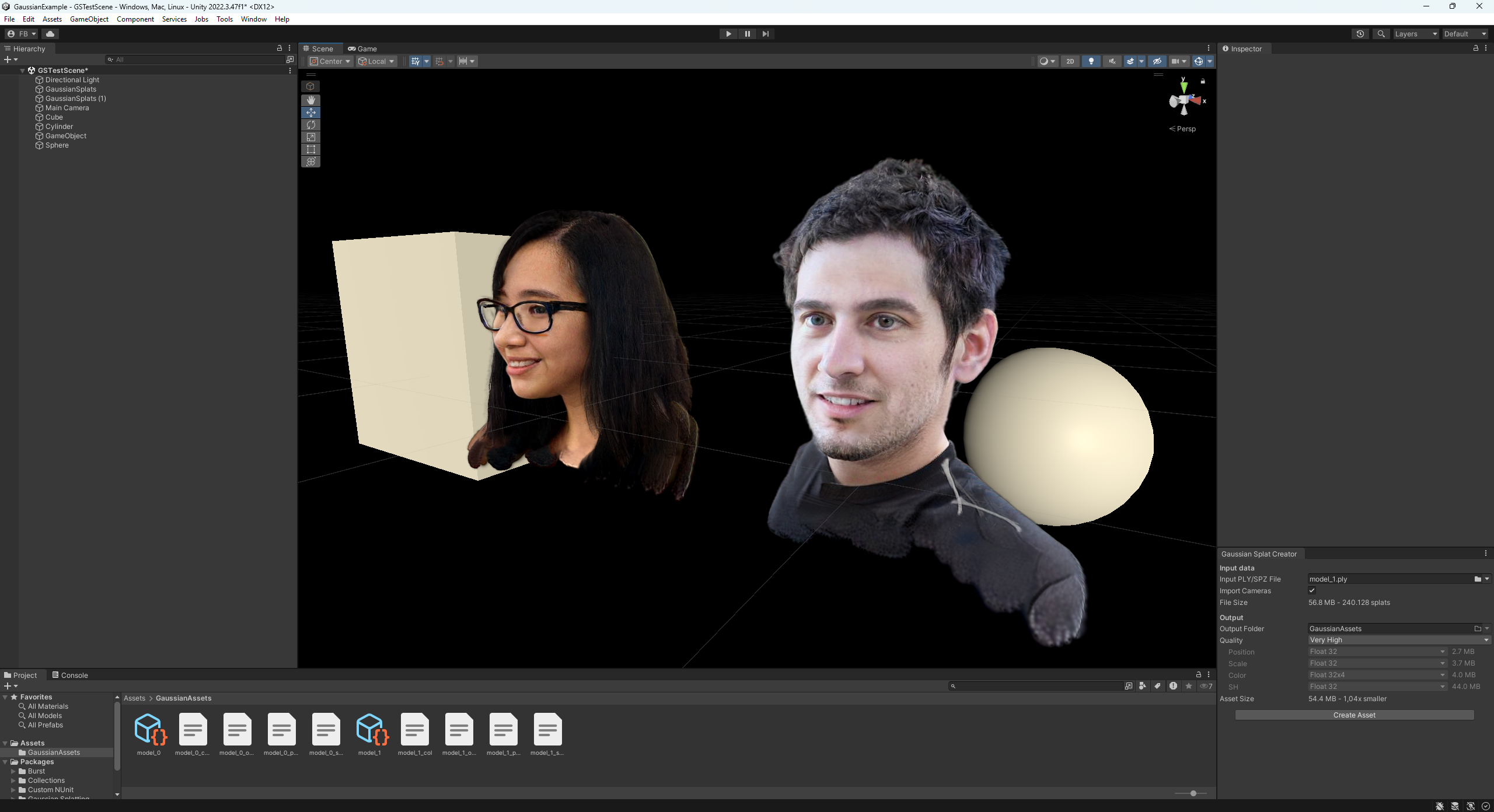}
    \caption{Our faces imported into unity using the Unity Gaussian splatting plugin by Aras \mbox{Pranckevičius}: \url{https://github.com/aras-p/UnityGaussianSplatting}}
    \label{fig:unity}
\end{figure}

In \autoref{fig:unity}, we demonstrate that our 3D heads can be imported into explicit 3D environments, such as Unity. Using the 3DGS plugin by Aras \mbox{Pranckevičius} (\url{https://github.com/aras-p/UnityGaussianSplatting}), the 3D faces can simply be placed into 3D scenes and interact with other 3D objects. Combining this with the inversion capabilities of our models, we can now create realistic 3D assets just from a single 2D image.

\newpage
\section*{NeurIPS Paper Checklist}

\begin{enumerate}

\item {\bf Claims}
    \item[] Question: Do the main claims made in the abstract and introduction accurately reflect the paper's contributions and scope?
    \item[] Answer: \answerYes{} 
    \item[] Justification: We give a brief overview of our work in the abstract and list our main contributions again in the introduction. 

\item {\bf Limitations}
    \item[] Question: Does the paper discuss the limitations of the work performed by the authors?
    \item[] Answer: \answerYes{} 
    \item[] Justification: We discuss the limitations for both our method and our curated dataset in more detail in the appendix, additionally showing failure cases.

\item {\bf Theory assumptions and proofs}
    \item[] Question: For each theoretical result, does the paper provide the full set of assumptions and a complete (and correct) proof?
    \item[] Answer: \answerNA{} 
    \item[] Justification: No theoretical results or proofs are made in our paper.

    \item {\bf Experimental result reproducibility}
    \item[] Question: Does the paper fully disclose all the information needed to reproduce the main experimental results of the paper to the extent that it affects the main claims and/or conclusions of the paper (regardless of whether the code and data are provided or not)?
    \item[] Answer: \answerYes{} 
    \item[] Justification:  Next to the details in our main paper, we provide more detailed information about specific training settings in the appendix.

\item {\bf Open access to data and code}
    \item[] Question: Does the paper provide open access to the data and code, with sufficient instructions to faithfully reproduce the main experimental results, as described in supplemental material?
    \item[] Answer: \answerYes{} 
    \item[] Justification: The training data that we use is publicly available and the code will be published along with detailed instructions how to reproduce our results. We also provide the code to the reviewers.

\item {\bf Experimental setting/details}
    \item[] Question: Does the paper specify all the training and test details (e.g., data splits, hyperparameters, how they were chosen, type of optimizer, etc.) necessary to understand the results?
    \item[] Answer: \answerYes{} 
    \item[] Justification: We report our training details in the appendix.

\item {\bf Experiment statistical significance}
    \item[] Question: Does the paper report error bars suitably and correctly defined or other appropriate information about the statistical significance of the experiments?
    \item[] Answer: \answerNo{} 
    \item[] Justification: Training our models and the models of related work takes up a lot of time and resources, not allowing us to run them multiple times.

\item {\bf Experiments compute resources}
    \item[] Question: For each experiment, does the paper provide sufficient information on the computer resources (type of compute workers, memory, time of execution) needed to reproduce the experiments?
    \item[] Answer: \answerYes{} 
    \item[] Justification: We report memory consumption and training time in \ref{sec:exp}.
    
\item {\bf Code of ethics}
    \item[] Question: Does the research conducted in the paper conform, in every respect, with the NeurIPS Code of Ethics \url{https://neurips.cc/public/EthicsGuidelines}?
    \item[] Answer: \answerYes{} 
    \item[] Justification: We review the NeurIPS code of Ethics and ensure that there is no conflict with our work.

\item {\bf Broader impacts}
    \item[] Question: Does the paper discuss both potential positive societal impacts and negative societal impacts of the work performed?
    \item[] Answer: \answerYes{} 
    \item[] Justification: We provide a short section about the broader impact of our work in the appendix.
    
\item {\bf Safeguards}
    \item[] Question: Does the paper describe safeguards that have been put in place for responsible release of data or models that have a high risk for misuse (e.g., pretrained language models, image generators, or scraped datasets)?
    \item[] Answer: \answerNA{} 
    \item[] Justification: Our paper does not pose such risks as we do not use any additional data other than publicly available image datasets.

\item {\bf Licenses for existing assets}
    \item[] Question: Are the creators or original owners of assets (e.g., code, data, models), used in the paper, properly credited and are the license and terms of use explicitly mentioned and properly respected?
    \item[] Answer: \answerYes{} 
    \item[] Justification: We only use FFHQ which was published along with the StyleGAN paper which is cited. FFHQ has a CC BY-NC-SA 4.0 license.

\item {\bf New assets}
    \item[] Question: Are new assets introduced in the paper well documented and is the documentation provided alongside the assets?
    \item[] Answer: \answerNA{}. 
    \item[] Justification: We do not release new assets. We only change the pre-processing of  existing public datasets.

\item {\bf Crowdsourcing and research with human subjects}
    \item[] Question: For crowdsourcing experiments and research with human subjects, does the paper include the full text of instructions given to participants and screenshots, if applicable, as well as details about compensation (if any)? 
    \item[] Answer: \answerNA{} 
    \item[] Justification: Our paper does not involve crowdsourcing nor research with human subjects.

\item {\bf Institutional review board (IRB) approvals or equivalent for research with human subjects}
    \item[] Question: Does the paper describe potential risks incurred by study participants, whether such risks were disclosed to the subjects, and whether Institutional Review Board (IRB) approvals (or an equivalent approval/review based on the requirements of your country or institution) were obtained?
    \item[] Answer: \answerNA{} 
    \item[] Justification: Our paper does not involve crowdsourcing nor research with human subjects.

\item {\bf Declaration of LLM usage}
    \item[] Question: Does the paper describe the usage of LLMs if it is an important, original, or non-standard component of the core methods in this research? Note that if the LLM is used only for writing, editing, or formatting purposes and does not impact the core methodology, scientific rigorousness, or originality of the research, declaration is not required.
    \item[] Answer: \answerNo{} 
    \item[] Justification: We exclusively use LLM for writing, editing, or formatting.

\end{enumerate}

\end{document}